\documentclass{article}
\usepackage{spconf,amsmath,graphicx,amssymb}
\usepackage{booktabs}
\usepackage{multirow}
\title{Transferring speech-generic and depression-specific knowledge for Alzheimer's disease detection}
\name{Ziyun Cui$^{1,\dagger}$, Wen Wu$^{2,\dagger}$\thanks{$\dagger$ Co-first author. $*$ Corresponding author.\\This work was supported by the OPPO Research Fund.\\Wen Wu is supported by a Cambridge International Scholarship from the Cambridge Trust.}, Wei-Qiang Zhang$^{1}$, Ji Wu$^{1}$, Chao Zhang$^{1,*}$}
\address{$^{1}$Department of Electronic Engineering, Tsinghua University, Beijing, China \\
    $^{2}$Department of Engineering, University of Cambridge, Trumpington St., Cambridge, UK \\
    \small{\texttt{cuizy20@mails.tsinghua.edu.cn, ww368@eng.cam.ac.uk, cz277@tsinghua.edu.cn}}
}

\copyrightnotice{979-8-3503-0689-7/23/\$31.00~\copyright2023 IEEE}

\begin{document}
%
\maketitle
\begin{abstract}
The detection of Alzheimer's disease (AD) from spontaneous speech has attracted increasing attention while the sparsity of training data remains an important issue. This paper handles the issue by knowledge transfer, specifically from both speech-generic and depression-specific knowledge. The paper first studies sequential knowledge transfer from generic foundation models pretrained on large amounts of speech and text data. A block-wise analysis is performed for AD diagnosis based on the representations extracted from different intermediate blocks of different foundation models. Apart from the knowledge from speech-generic representations, this paper also proposes to simultaneously transfer the knowledge from a speech depression detection task based on the high comorbidity rates of depression and AD.
A parallel knowledge transfer framework is studied that jointly learns the information shared between these two tasks. Experimental results show that the proposed method improves AD and depression detection, and produces a state-of-the-art F1 score of 0.928 for AD diagnosis on the commonly used ADReSSo dataset. 

\end{abstract}
\section{Introduction}
\label{sec:intro}

Alzheimer's disease (AD) is a neurodegenerative disease entailing a long-term and gradual decrease of cognitive functioning~\cite{deture2019neuropathological}. Detection of AD has attracted extensive attention worldwide~\cite{crous2017alzheimer} and the need for more cost-effective and scalable AD detection methods motivates research in automated AD detection from spontaneous speech~\cite{luz21_interspeech}. Speech produced by AD patients has some special acoustic and linguistic patterns. For example, people with AD tend to use more ``empty'' words and fewer information-bearing nouns and especially verbs, and their discourse appears to be disorganized and produces more pauses and hesitations~\cite{yuan2020disfluencies, szatloczki2015speaking}. Speech is an effective method of screening AD while automatic AD detection suffers from a lack of data due to difficulties in data collection, high labelling costs, privacy issues, \textit{etc.} This paper investigates knowledge transfer to handle the data sparsity issue of automatic AD diagnosis from spontaneous speech.

Foundation models are sizeable neural network models trained on a large amount of data at scale, which encode generic knowledge and can be adapted to a wide range of downstream tasks~\cite{yang2021superb,wu2023self}. Transferring generic knowledge from the foundation models can compensate for the lack of data for the downstream tasks.  Foundation models have produced superior performance on many speech processing tasks such as automatic speech recognition~\cite{chang2021exploration}, emotion recognition~\cite{morais2022speech}, and speaker verification~\cite{chen2022large} while whether such speech-generic representations contain transferable knowledge for Alzheimer's disease diagnosis is still understudied. 

Meanwhile, clinical studies have shown that AD is closely related to depression that over 80\% of patients with AD develop non-cognitive neuropsychiatric symptoms during the course of their illness, among which depression is the most frequent of such comorbidities, affecting up to 50\% of AD patients~\cite{lyketsos2003diagnosis}. In addition, a history of depression may confer an increased risk for later developing AD, even in families where first depression symptoms occurred more than 25 years before the onset of AD~\cite{green2003depression, ownby2006depression}. Depression and AD share genetic bases, as revealed by a study based on a comprehensive characterization from the behavioural to transcriptomic level~\cite{martin2021comorbidity}. 
This indicates that depression-specific representations may contain useful knowledge for AD diagnosis.

This paper investigates the use of speech-generic and depression-specific knowledge for AD diagnosis. A sequential knowledge transfer framework is developed to leverage speech-generic knowledge from pretrained speech foundation models. A block-wise analysis is performed for AD diagnosis based on the representations from different intermediate blocks of multiple foundation models, which also provides insights into the effectiveness of speech-generic information from different sources for AD diagnosis. Moreover, a parallel knowledge transfer framework is proposed to investigate the usefulness of depression-specific information for AD diagnosis. The system simultaneously transfers the knowledge from a speech depression detection task to AD detection and jointly learns the information shared between them.

The rest of the paper is organised as follows. Section~\ref{sec: literature} summarises the related work on automatic AD detection.
Section~\ref{sec:method} introduces the proposed methods. The experimental setup is shown in Section \ref{sec: setup}, followed by the results of speech-generic and depression-specific knowledge transfer in Section~\ref{sec: foundation} and \ref{sec: dep} respectively. We conclude in Section \ref{sec:conclusion}.

\section{Related work}
\label{sec: literature}


Traditional clinical AD diagnosis is usually conducted through clinical assessment, cognitive deficit testing and neuroimaging. Given the growing AD population, more cost-effective automatic diagnosis of AD has attracted much attention in recent years~\cite{nestor2004advances, noor2020application}. Research has shown that AD is detectable through audio features including hand-crafted features such as short-time energy, spectral centroid~\cite{lopez2012new}, fluency~\cite{konig2015automatic}, spectrograms~\cite{bertini2022automatic,chen2021automatic} and also in deep-learning-model-extracted features~\cite{koo2020exploiting,zhu2021wavbert,chen2023cross}. Text information has also been investigated for AD detection such as GloVe embeddings \cite{rohanian2021alzheimer} and representations extracted from pretrained large language models~\cite{agbavor2022predicting}. Furthermore, various methods have been proposed to fuse audio and text modalities~\cite{rohanian2021alzheimer,haulcy2021classifying}, which further improved the performance.
Apart from input features, research has studied various classifiers for automatic AD detection including statistical models such as support vector machines~\cite{lopez2012new, konig2015automatic, agbavor2022predicting}, logistic regression~\cite{chen2021automatic, agbavor2022predicting}, random forest~\cite{agbavor2022predicting} and deep learning models such as  bidirectional recurrent neural networks~\cite{koo2020exploiting} and auto-encoders~\cite{bertini2022automatic}.


Despite the extensive research on AD detection, few studies investigate the connection between AD and depression. Villatoro \textit{et al.}~\cite{villatoro2021late} proposed a late fusion method to combine lexicon and acoustic features extracted from convolutional neural networks (CNNs). The method was applied to detect AD and depression separately and did not combine them together. Perez \textit{et al.}~\cite{perez2023transferring} leveraged quantified 
emotion knowledge to detect depression in AD patients, which required every data sample to have both AD and depression annotations and limited the number of samples available for training. 
To the best of our knowledge, this paper is the first research on cross-corpus knowledge transfer from depression detection to AD detection.

\section{Proposed method}
\label{sec:method}

\subsection{System structure}
\begin{figure}[tb]
\centering
\includegraphics[width=0.7\linewidth]{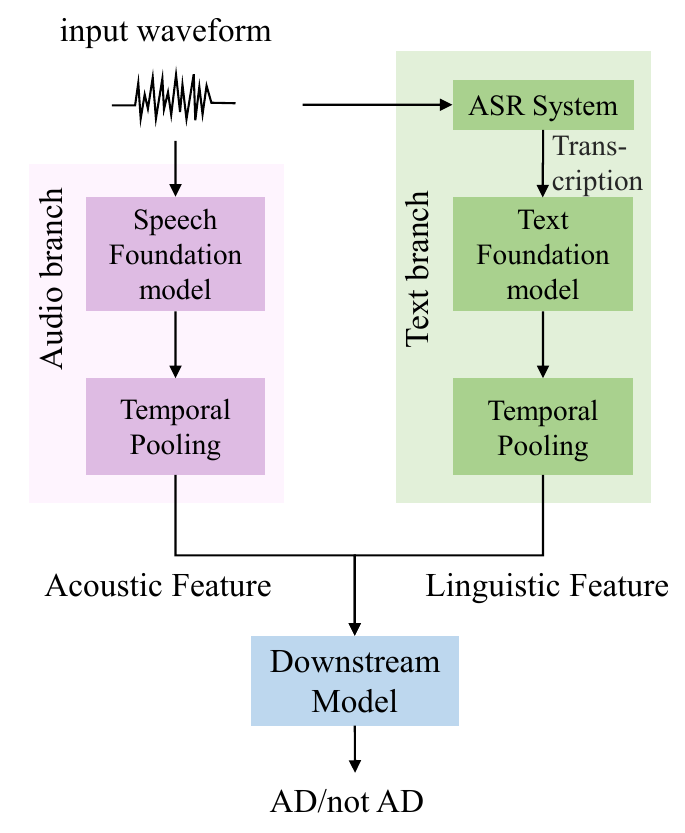}
\caption{Overall structure of the proposed system.}
\label{fig: overall}
\end{figure}

The structure of the proposed system is shown in Fig.~\ref{fig: overall}, which consists of two branches. AD detection is conducted on a dialogue $\{\mathbf{x}_1, \mathbf{x}_2,\ldots, \mathbf{x}_N\}$ where $N$ is the number of utterances in the dialogue. For each utterance  $\mathbf{x}_n$, taking the raw waveform of the utterance, the audio branch extracts acoustic features by a pretrained speech foundation model. The acoustic features have a shape of $(T_n^\text{acou}, D^\text{acou})$ where $T_n^\text{acou}$ is the number of audio frames of $\mathbf{x}_n$ and $D^\text{acou}$ is the dimension of hidden states of the speech foundation model. In the text branch, an automatic speech recognition (ASR) model is used to transcribe the input speech to text. The transcription is then encoded by a text foundation model, which produces text feature of shape $(T_n^\text{text}, D^\text{text})$, where $T_n^\text{text}$ is the number of tokens in the text transcriptions and $D^\text{text}$ is the hidden state dimension of the text foundation model. An utterance-level temporal pooling is applied to both acoustic and text features, resulting in a $D^\text{text}$ dimensional text feature and a $D^\text{acou}$ dimensional acoustic feature for each utterance. In a dialogue with $N$ utterances, a text feature sequence of shape $(N, D^\text{text})$ and an acoustic feature sequence of shape $(N, D^\text{acou})$ are concatenated and fed into the downstream block to predict whether the speaker has AD or not.

\subsubsection{Speech foundation model}
Three speech foundation models are investigated: wav2vec 2.0 (W2V2) \cite{baevski2020wav2vec}\footnote{https://huggingface.co/facebook/wav2vec2-base}, HuBERT \cite{hsu2021hubert}\footnote{https://huggingface.co/facebook/hubert-base-ls960}, WavLM \cite{chen2022wavlm}\footnote{https://huggingface.co/microsoft/wavlm-base-plus}. W2V2 is pretrained by a contrastive task defined over a quantization of the latent representations, where the latent space is jointly learnt during training. HuBERT initialises the latent space by an offline clustering step which serves as aligned target labels for a BERT-like prediction loss. WavLM is built based on the HuBERT framework with an emphasis on both spoken content modelling and speaker identity preservation.  In this study, the ``BASE'' version is used for W2V2 and HuBERT and the ``BASE+'' version is used for WavLM,  which all consist of a CNN feature extractor, followed by 12 Transformer encoder blocks with a hidden dimension of 768. 


\subsubsection{ASR system and text foundation model}
A pretrained Whisper model~\cite{radford2022robust}\footnote{https://huggingface.co/openai/whisper-small} is used to transcribe the input utterance.  Whisper is a series of state-of-the-art ASR models based on the encoder-decoder Transformer architecture and pretrained on 680,000 hours of multi-lingual data by multi-task supervision. ``SMALL'' version is used in this paper which contains 12 Transformer encoder layers.

The transcriptions are encoded by a pretrained BERT model~\cite{devlin2018bert}\footnote{https://huggingface.co/bert-base-uncased}. The ``BASE-uncased'' version is used in the paper, which contains 12 bidirectional Transformer encoders.





\subsubsection{Downstream AD detection block}
The downstream AD detection block consists of two 128-d Transformer encoders with four attention heads each, to capture the contextual dependencies within the dialogue, followed by two fully connected (FC) layers.


\subsection{Speech-generic knowledge transfer}

It has been shown that block-wise evolution of the representations extracted from different Transformer blocks of the speech foundation models tends to follow an acoustic-linguistic hierarchy~\cite{pasad2021layer,pasad2023comparative}. In general, the shallowest layers encode acoustic features, followed by phonetic, word identity, and word meaning information~\cite{pasad2021layer} while which specific layer does useful phone/word information tends to concentrate depending on the pretraining process of different models. 
In this paper, block-wise analysis is performed to transfer different levels of speech-generic knowledge encoded by the pretrained foundation models to AD diagnosis. The analysis can also provide insight into what kind of information is more relevant to AD diagnosis.

\begin{figure}[tb]
\centering
\includegraphics[width=\linewidth]{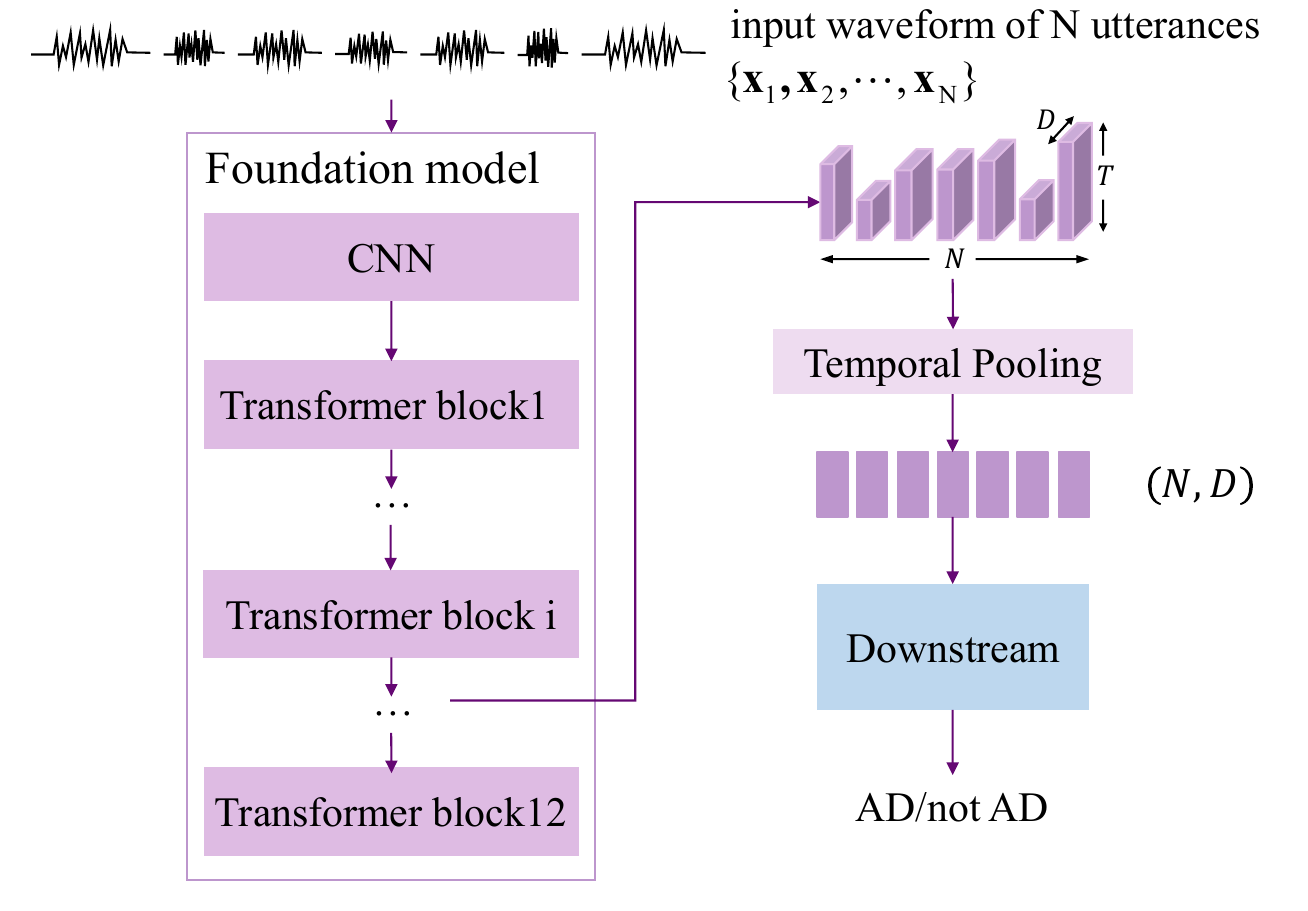}
\caption{The structure of block-wise analysis of speech foundation models.}
\label{fig: block-wise}
\end{figure}

The structure of the block-wise analysis is illustrated in Fig.~\ref{fig: block-wise}. The outputs from each Transformer block of the speech foundation model are extracted and temporal pooling is applied along dimension $T$. The downstream model takes the pooled features as input and is trained for binary AD classification through cross-entropy loss.

\begin{table*}[t]
\centering
\resizebox{\textwidth}{!}{%
\begin{tabular}{c|ccc|ccc|ccc|ccc}
\toprule
 & \multicolumn{3}{c|}{WavLM$_\text{PT}$} & \multicolumn{3}{c|}{HuBERT$_\text{PT}$} & \multicolumn{3}{c|}{W2V2$_\text{PT}$} & \multicolumn{3}{c}{WavLM$_\text{AER}$} \\ \midrule
Layer & F1-avg & F1-max & F1-std & F1-avg & F1-max & F1-std & F1-avg & F1-max & F1-std & F1-avg & F1-max & F1-std \\
1 & 0.735 & 0.761 & 0.015 & 0.644 & 0.667 & 0.019 & 0.662 & 0.692 & 0.019 & 0.723 & 0.750 & 0.022 \\
3 & 0.738 & 0.789 & 0.039 & 0.680 & 0.744 & 0.043 & 0.710 & 0.725 & 0.026 & 0.695 & 0.709 & 0.015 \\
5 & 0.746 & 0.785 & 0.025 & 0.660 & 0.703 & 0.025 & 0.714 & 0.735 & 0.008 & \textbf{0.761} & 0.765 & 0.004 \\
7 & 0.729 & 0.773 & 0.037 & 0.703 & 0.763 & 0.038 & 0.703 & 0.722 & 0.023 & 0.743 & 0.750 & 0.005 \\
9 & 0.742 & 0.773 & 0.027 & \textbf{0.750} & \textbf{0.789} & 0.049 & \textbf{0.723} & \textbf{0.765} & 0.022 & 0.743 & 0.757 & 0.013 \\
11 & \textbf{0.763} & \textbf{0.805} & 0.037 & 0.717 & 0.734 & 0.014 & 0.686 & 0.725 & 0.027 & 0.749 & \textbf{0.784} & 0.023 \\
\midrule
Weighted & 0.750 & 0.769 & 0.019 & 0.698 & 0.732 & 0.030 & 0.718 & 0.754 & 0.024 & 0.755 & 0.773 & 0.014 \\ \bottomrule
\end{tabular}%
}
\caption{The F1 results of block-wise analysis of different speech foundation models on AD detection task. Only AD data is used.
The best values in each column are shown in bold.} 
\label{tab:block-wise}
\end{table*}

\subsection{Depression-specific knowledge transfer}

\begin{figure}[tb]

\begin{minipage}[b]{1.0\linewidth}
  \centering
  \centerline{\includegraphics[width=8.5cm]{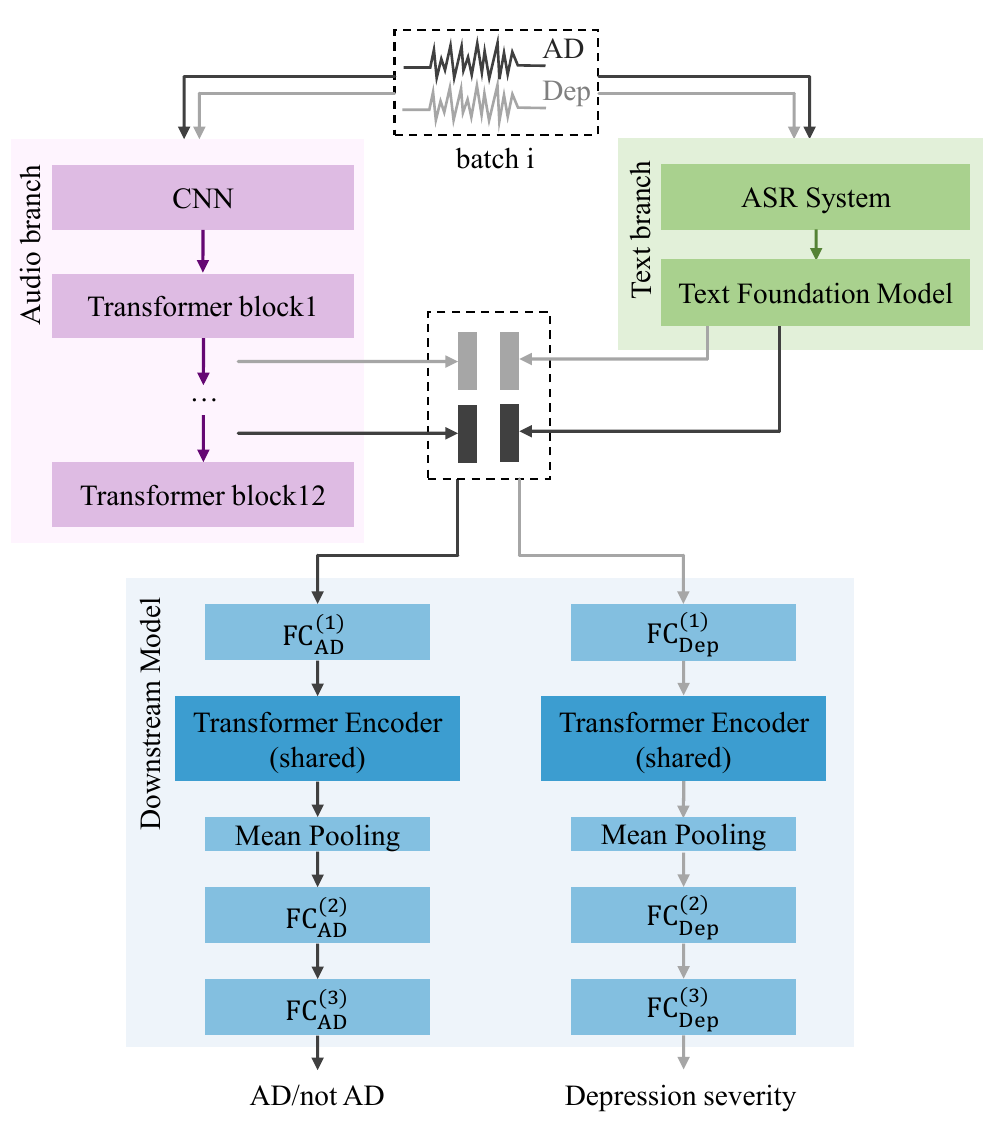}}
\end{minipage}
\caption{The structure of the downstream model for knowledge transfer learning.}
\label{fig:dep}
\end{figure}

In light of the close connection between AD and depression, this paper investigates whether depression-specific representations include valuable evidence for AD diagnosis. A parallel knowledge transfer framework is proposed where the model is jointly trained for depression and AD detection on separate datasets. 
The structure of the proposed system is shown in Fig.~\ref{fig:dep}. Approximately a balanced amount of AD and depression data are sampled within each training batch. AD and depression data are separately encoded by the same upstream model. The downstream model contains a Transformer block shared by two tasks to capture the correlation between them, and each task has its own dimension reduction and output layers to learn the task-specific knowledge. Loss functions are computed separately for the two tasks and are added up to form an overall loss to compute the gradients in backward propagation, to achieve simultaneous parameter updates for both tasks and prevent the model from catastrophic forgetting.

The proposed system allows the integration of both AD and depression data by leveraging shared encoder blocks while maintaining separate processing streams, which can not only capture the unique characteristics of AD and depression but also learn the shared information between them.

\section{Experimental setup}
\label{sec: setup}
\subsection{Datasets}
\label{ssec:Dataset}

The ADReSSo dataset~\cite{luz21_interspeech} is used in this paper, which contains recordings of picture descriptions produced by cognitively normal subjects and patients with AD, who were asked to describe the Cookie Theft picture from the Boston Diagnostic Aphasia Examination. ADReSSo contains 237 recordings, with a total length of 304 minutes. 115 out of 237 are classified as AD. The standard split of train/test provided by the ADReSSo corpus is used. 20\% of the training data was further set aside for validation.


The dataset used for depression is the DAIC-WOZ~\cite{gratch2014distress} depression dataset, which contains interviews between participants and an interviewer. DAIC-WOZ corpus consists of 189 interviews, with a total length of 50+ hours. Participants of 56 interviews are depressed. The standard split of train/validation/test provided by the dataset was used.


\subsection{Data augmentation}
Due to the sparsity of training data, sub-dialogue shuffling~\cite{wu2023self} was applied to augment the training data. For a dialogue consisting of $N$ utterances $\{x_1, x_2,\ldots, x_N\}$, where the average value of $N$ is 18, we randomly sampled sub-dialogue $\{x_a, x_{a+1}, ..., x_b\}$ with random starting index $a$ and ending index $b$ where $b-a$ was randomized within a range of 0.5$N$ to $N$. For each training dialogue in AD dataset, 50 sub-dialogues were sampled, trading off between model performance and computational resources. The amounts between AD dataset and depression dataset were balanced by adjusting the number of sub-dialogues.


\subsection{Implementation details}
The model was implemented in PyTorch using the Speechbrain toolkit~\cite{speechbrain}. Parameters of the pretrained foundation models were frozen and only the downstream block was updated. A Dropout rate of 0.2 was applied to the Transformer encoders in the downstream model. The system was trained using the Adam optimiser with a weight decay of 5$\times10^{-3}$. The linear scheduler was applied with the learning rate linearly decreasing from 4$\times10^{-5}$ to 1$\times10^{-5}$. For each random seed, the model was trained for 30 epochs and the model with the best validation performance was selected for testing. 

The classification performance of AD task is evaluated by the F1 score. The model was initialised and trained for 5 different random seeds. Both the highest (F1-max) and the average (F1-avg) values are reported, along with the standard deviation (F1-std). Because of the imbalance of depression data, the performance of depression detection is evaluated using the root mean square error (RMSE) between the prediction and the ground truth.

Reference transcriptions are not provided in the ADReSSo dataset. ASR performance of the Whisper model is evaluated on ADReSS dateset~\cite{luz2020alzheimer} instead to provide a reference for the readers. The ADReSS dataset is a prior version of the ADReSSo dataset that includes reference transcriptions paired with audio. The Whisper model has a word error rate (WER) of 44.0\% on the ADReSS dataset. For comparison, W2V2 finetuned on 960 hours of Librispeech~\cite{panayotov2015librispeech}\footnote{https://huggingface.co/facebook/wav2vec2-base-960h} has a WER of 59.0\%.

When doing parallel knowledge transfer, cross-entropy loss was used for AD detection and MSE loss was used for depression severity prediction. The total loss is calculated by $\text{loss}_\text{AD}+\lambda\text{loss}_\text{Dep}$, in which the coefficient $\lambda$ is set to 0.1 to balance the dynamic range of the losses.




\section{Experimental Results of Transferring Speech-generic knowledge}
\label{sec: foundation}

\subsection{Block-wise analysis of speech foundation models}
\label{sec:Block-wise}
\begin{figure}[tb]
  \centering
  \includegraphics[width=0.85\linewidth]{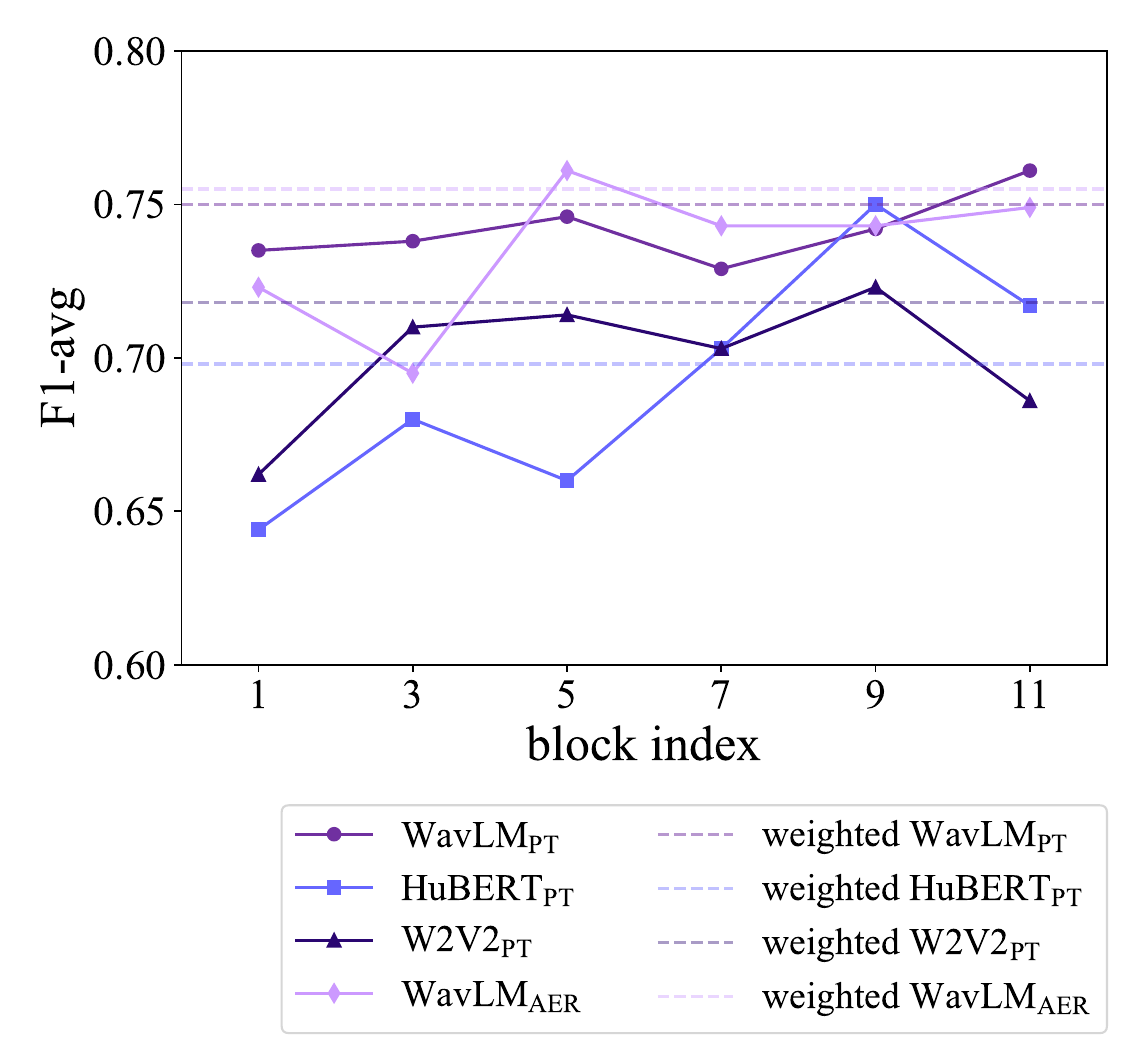}
\vspace{-1ex}
\caption{Summary of block-wise analysis of AD diagnosis. 
}
\label{fig2}
\end{figure}

The AD diagnosis results using output from different intermediate blocks of the pretrained (PT) W2V2 (W2V2$_\text{PT}$), HuBERT (HuBERT$_\text{PT}$), and WavLM (WavLM$_\text{PT}$) are presented in Table~\ref{tab:block-wise}. The F1-avg score of the models is compared in Fig.~\ref{fig2}. Among the three pretrained models, WavLM$_\text{PT}$ achieves the highest F1 score. The trend of the WavLM$_\text{PT}$ model has a peak at 11$^\text{th}$ block while W2V2$_\text{PT}$ and HuBERT$_\text{PT}$ get the highest F1-avg score at 9$^\text{th}$ block. According to previous findings~\cite{pasad2023comparative}, these blocks contain more phonetic and word-level information, indicating that phonetic and word-level information can be important in the diagnosis of AD.


The results of using a weighted combination of all 12 blocks are shown in the last row of Table~\ref{tab:block-wise}, where the weights were trained together with the downstream model. It is found that the combination of output from different intermediate blocks does not necessarily yield superior results compared with the best-performing single block, which aligns with the findings reported in \cite{pasad2023comparative}.

Emotion information has been shown useful for detecting depression~\cite{wu2023self,li2023leveraging}. We then investigate whether it is also helpful for AD detection. The WavLM$_\text{PT}$ model was finetuned on 110 hours of MSP-Podcast dataset~\cite{lotfian2017building} for automatic emotion recognition (AER). The finetuned model is denoted as WavLM$_\text{AER}$. As shown in Fig.~\ref{fig2}, the trend of WavLM$_\text{AER}$ has an earlier peak at the 5$^\text{th}$ block and does not surpass the best results of the WavLM$_\text{PT}$ model, which indicates that emotion information is not as useful for AD as for depression.


\subsection{Incorporating text information}
\label{sec:transcriptions}

\begin{table}[tb]
    \centering
    \begin{tabular}{c|ccc}
    \toprule
    \multicolumn{4}{c}{Whisper}\\
    \midrule
Hidden states & F1-avg & F1-max & F1-std\\
\midrule
1 & 0.638 & 0.658 & 0.013\\
3 & 0.620 & 0.667 & 0.030 \\
5 & 0.685 & 0.734 & 0.049 \\
7 & \textbf{0.717} & \textbf{0.730} & 0.014\\
9 & \textbf{0.717} & 0.727 & 0.016 \\
11 & 0.705 & 0.723 & 0.012 \\ 
\midrule
BERT$_\text{ASR}$ &  \textbf{0.826} & \textbf{0.880} & 0.049 \\ 
\bottomrule
    \end{tabular}
    \caption{Comparison of hidden states extracted from different Transformer blocks of the Whisper model and encoding the transcriptions obtained from the Whisper model by the BERT model on AD detection. }
    \label{tab: text}
\end{table}
As discussed above, linguistic information is effective for AD detection. This section investigates incorporating linguistic information in AD detection. Two ways of extracting linguistic features were compared: (i) extracting hidden states from different intermediate Transformer blocks of the Whisper model; (ii) obtaining transcriptions from the output of the Whisper model and encoding the transcriptions by a BERT model. The results are shown in Table~\ref{tab: text}. The BERT embeddings extracted using the ASR transcriptions (denoted as BERT$_\text{ASR}$) produced better performance than directly using hidden states of the ASR model. 

The BERT embeddings were then concatenated with the output of the 11$^\text{th}$ block of the WavLM models to investigate the fusion of two modalities. As shown in Table~\ref{tab: MMSE}, incorporating text information yields a notable improvement in the AD diagnosis task for both WavLM$_\text{PT}$ and WavLM$_\text{AER}$. It's worth noticing that although WavLM$_\text{PT}$ alone gives better results than WavLM$_\text{AER}$, they produced comparable performance when combined with text information. The system based on WavLM$_\text{AER}$ even slightly outperforms the system based on WavLM$_\text{PT}$. A possible explanation is that combining text information with emotion information can better exploit the complementarity between semantic and non-semantic information.

\begin{table}[bt]
\centering
\begin{tabular}{cc|ccc}
\toprule
Speech & Text  & F1-avg & F1-max & F1-std \\ \midrule
WavLM$_\text{PT}$  & None & 0.763 & 0.805 & 0.037 \\
WavLM$_\text{AER}$ & None & 0.749 & 0.784 & 0.023 \\
None & BERT$_\text{ASR}$ &  0.826 & 0.880 & 0.049 \\ 
\midrule
WavLM$_\text{PT}$ & BERT$_\text{ASR}$ &  0.854 & 0.880 & 0.021 \\
WavLM$_\text{AER}$ & BERT$_\text{ASR}$ & \textbf{0.857} & \textbf{0.895} & 0.035 \\ \bottomrule
\end{tabular}%
\caption{AD diagnosis results with acoustic and text features.}
\label{tab: MMSE}
\end{table}

\section{Experimental Results of Transferring depression-specific knowledge}
\label{sec: dep}

\begin{table}[t]
\centering
\resizebox{\columnwidth}{!}{%
\begin{tabular}{cc|cc|cc}
\toprule
\multirow{2}{1em}{AD} & \multirow{2}{1.5em}{Dep} & \multicolumn{2}{c|}{AD} & \multicolumn{2}{c}{Depression} \\
&  & F1-avg & F1-max & RMSE-avg & RMSE-min \\ \midrule
\checkmark & & 0.857 & 0.895 & / & / \\ 
& \checkmark & / & / & 6.91 & 6.75 \\ 
\midrule
\checkmark & \checkmark & \textbf{0.894} & \textbf{0.928} & \textbf{6.01} & \textbf{5.58} \\ 
\bottomrule
\end{tabular}%
}
\caption{The results of depression-specific knowledge transfer. Depression data from DAIC-WOZ dataset is used.}
\label{tab:depression-daic}
\end{table}

This section investigates depression-specific knowledge transfer by jointly training the model on depression and AD data. Hidden states of the 11$^\text{th}$ Transformer block of the WavLM model were used for AD acoustic feature. Previous study~\cite{wu2023self} suggests that 8$^\text{th}$ - 10$^\text{th}$ blocks contain the most effective information for depression detection. Therefore, embeddings extracted from the 9$^\text{th}$ block were used for the depression feature. Depression data from the DAIC-WOZ dataset was used, results shown in Table~\ref{tab:depression-daic}. Compared to the system trained solely on AD data, DAIC-WOZ leads to improvement in AD detection, with an increasing from 0.857 to 0.894 on average F1 score. Furthermore, improvement is also observed in depression detection performance. RMSE of the depression severity prediction of the DAIC-WOZ dataset drops from 6.91 to 6.01 on average and from 6.75 to 5.58 for the best-performing seed. 

Compared with sequential knowledge transfer where the pretrained model is finetuned for another task, the benefit of the proposed parallel transfer framework is that the model jointly learns about both tasks. In sequential knowledge transfer, the model gradually forgets the knowledge about the task used for pretraining when it is finetuned on another task while in the proposed framework, the model is trained on both tasks simultaneously which helps the model learn the cross-domain information between the tasks as well as retain the domain-specific knowledge of each tasks.



Results show that the parallel knowledge transfer between depression detection and AD detection improves the performance for both, which verifies the underlying connection between AD and depression. AD is associated with intact experience but the abnormal expression of emotion, while depression involves difficulties in emotional regulation caused by cognitive biases and deficits~\cite{henry2009emotion, joormann2014cognitive}. Besides, the combination of corpora from related tasks increases the amount of available training data which provides a more comprehensive and diverse representation of speech patterns and can potentially compensate for the data sparsity issue of each individual task. The model can then benefit from a richer set of examples and can better capture the shared characteristics between the tasks. In addition, the success of parallel knowledge transfer of AD and depression inspires the further development of more comprehensive systems which are capable of diagnosing various types of mental and cognitive disorders.

Finally, we cross-compare our results with the literature. As shown in Table~\ref{tab: cross compare}, our proposed approach gives state-of-the-art AD detection performance in terms of F1-max values.

\begin{table}[tb]
\centering
\begin{tabular}{c|c}
\toprule
Paper & F1-max\\
\midrule
Baseline~\cite{luz21_interspeech}& 0.789\\
Agbavor \textit{et al.}~\cite{agbavor2022predicting} & 0.829 \\
Pappagari \textit{et al.}~\cite{pappagari2021automatic} & 0.860\\
Deng \textit{et al.}~\cite{deng2022alzheimer} & 0.873 \\
Chen \textit{et al.}~\cite{chen2021automatic} & 0.889\\
Priyadarshinee \textit{et al.}~\cite{priyadarshinee2023alzheimer}& 0.894 \\
\midrule
Ours & \textbf{0.928} \\
\bottomrule
\end{tabular}%
\caption{Cross comparison of F1 score on ADReSSo test set.}
\label{tab: cross compare}
\end{table}

\section{Conclusion}
\label{sec:conclusion}
This paper investigates speech-generic and depression-specific knowledge transfer for Alzheimer’s disease detection. 
The use of speech-generic knowledge is studied first based on a block-wise analysis with three speech foundation models, which found the importance of phonetic information and word-level information in AD diagnosis. 
An end-to-end system structure is proposed that captures the cross-domain information between the AD and depression detection tasks by sharing the encoder blocks. Separate processing streams are maintained for each task in the structure to retain their domain-specific knowledge.   
As a result, our proposed system achieves a state-of-the-art F1 score of 0.928 on the ADReSSo test set by transferring both speech-generic and depression-specific knowledge. 
Improvements are also observed over each individual task that verifies the connection between AD and depression.

\bibliographystyle{IEEEbib}
\bibliography{strings,refs}

\end{document}